\documentclass[11pt,a4paper]{article}%
\usepackage[T1]{fontenc}
\usepackage{libertine}
\usepackage[table]{xcolor}
\usepackage[american]{babel}
\usepackage{xfrac}
\usepackage{bbm}
\usepackage{authblk}
\usepackage{amsthm,amssymb}
\usepackage{bm}

\usepackage[format=plain,font=it]{caption}
\usepackage{graphicx}
\usepackage{amsmath}
\usepackage{amsthm}
\usepackage{booktabs}
\usepackage{algorithm}
\usepackage{algorithmic}
\usepackage[switch]{lineno}
\usepackage{pdflscape}
\usepackage{arydshln}
\usepackage{forest}
\usepackage{verbatim}

\usepackage{geometry}
\usepackage[round]{natbib} 
\usepackage{mathtools} 
\usepackage{booktabs} 
\usepackage{tikz} 
\usepackage{multirow}
\usepackage{listings}
\usepackage{float}
\usepackage{subfig}
\usepackage{amsmath}
\usepackage{amsfonts}
\usepackage{comment}
\usepackage{dsfont}

\newcommand{\indep}{\perp \!\!\! \perp}
\usepackage{array}
\newcolumntype{C}[1]{>{\centering\arraybackslash}m{#1}}
\newcolumntype{R}[1]{>{\raggedleft\arraybackslash}m{#1}}
\newcolumntype{L}[1]{>{\raggedright\arraybackslash}m{#1}}

\newcommand{\argmin}[1]{\underset{#1}{\mathrm{argmin}}}

\let\proglang=\textsf
\newcommand{\pkg}[1]{{\fontseries{m}\fontseries{b}\selectfont #1}}

\makeatletter
\newcommand\code{\bgroup\@makeother\_\@makeother\~\@makeother\$\@codex}
\def\@codex#1{{\normalfont\ttfamily\hyphenchar\font=-1 #1}\egroup}
\makeatother

\title{EquiPy: Sequential Fairness using \\Optimal Transport in \proglang{Python}\thanks{Agathe Fernandes Machado acknowledges that the project leading to this publication has received funding from OBVIA and \textit{Centre de Recherches Mathématiques} (CRM). Arthur Charpentier acknowledges funding from the SCOR Foundation for Science and the National Sciences and Engineering Research Council (NSERC) for funding (RGPIN-2019-07077).}} 

\usepackage{graphicx,pstricks,psfrag}
\definecolor{bleu}{RGB}{0,101,189}
\definecolor{vert}{HTML}{004D40}
\definecolor{rose}{HTML}{D81B60}
\definecolor{bleuTOL}{HTML}{332288}
\definecolor{wongBlack}{RGB}{0,0,0}
\definecolor{wongGold}{RGB}{230, 159, 0}
\definecolor{wongLightBlue}{RGB}{86, 180, 233}
\definecolor{wongGreen}{RGB}{0, 158, 115}
\definecolor{wongYellow}{RGB}{240, 228, 66}
\definecolor{wongBlue}{RGB}{0, 114, 178}
\definecolor{wongOrange}{RGB}{213, 94, 0}
\definecolor{wongPurple}{RGB}{204, 121, 167}
\definecolor{colUncalibrated}{RGB}{191, 191, 191}
\definecolor{colRecalibrated}{RGB}{197, 214, 231}

\definecolor{bleuTOL}{HTML}{332288}
\definecolor{vertTOL}{HTML}{117733}
\definecolor{vertClairTOL}{HTML}{44AA99}
\definecolor{bleuClairTOL}{HTML}{88CCEE}
\definecolor{sableTOL}{HTML}{DDCC77}
\definecolor{parmeTOL}{HTML}{CC6677}
\definecolor{magentaTOL}{HTML}{AA4499}
\definecolor{roseTOL}{HTML}{882255}

\definecolor{wongPurple}{RGB}{204, 121, 167}
\definecolor{wongLightBlue}{RGB}{86, 180, 233}
\definecolor{gris}{HTML}{A9A9A9}

\usepackage{hyperref}
\hypersetup{
    plainpages=false,
    bookmarksopen,
    bookmarksnumbered,
    unicode=true,          
    pdftoolbar=true,        
    pdfmenubar=true,        
    pdffitwindow=false,     
    pdfstartview={FitH},    
    pdftitle={EquiPy: Sequential Fairness using
Optimal Transport in Python},    
    pdfauthor={Fernandes Machado, Agathe and Grondin, Suzie and Ratz, Philipp and Charpentier, Arthur and Hu, François},     
    pdfkeywords={algorithmic fairness, optimal transport, sequential calibration, 
    Python package}, 
    pdfnewwindow=true,      
    colorlinks=true,       
    linkcolor=black,          
    citecolor={black},        
    filecolor={rose},      
    urlcolor={wongBlue},           
}

\usepackage{doi}
\author[1]{Agathe~Fernandes~Machado\thanks{Corresponding author: \href{mailto:fernandes_machado.agathe@courrier.uqam.ca}{fernandes\_machado.agathe@courrier.uqam.ca}}}
\author[2]{Suzie~Grondin}
\author[1]{Philipp~Ratz}
\author[1]{Arthur~Charpentier}
\author[3]{François~Hu}

\affil[1]{%
    \footnotesize Département de Mathématiques\\
    Université du Québec à Montréal\\
    Montréal, Québec, Canada
}
\affil[2]{%
    \footnotesize ENSAE Institut Polytechnique\\
    Paris, France
}
\affil[3]{%
    \footnotesize Milliman\\
    Paris, France
}

\usepackage[misc]{ifsym}
\makeatletter
\def\@fnsymbol#1{%
   \ifcase#1\or
   \TextOrMath ~ \dagger\or
   \TextOrMath {\footnotesize\Letter} \dagger\or
   \TextOrMath \textdaggerdbl \ddagger \or
   \TextOrMath \textsection  \mathsection\or
   \TextOrMath \textparagraph \mathparagraph\or
   \TextOrMath \textbardbl \|\or
   \TextOrMath {\textdagger\textdagger}{\dagger\dagger}\or
   \TextOrMath {\textdaggerdbl\textdaggerdbl}{\ddagger\ddagger}\else
   \@ctrerr \fi
}
\makeatother

\usepackage{fancyhdr} 
\newcommand{\authornames}{\footnotesize\textsc{Fernandes Machado, Grondin, Ratz, Charpentier, Hu}}
\usepackage{etoolbox}
\makeatletter
\patchcmd{\NAT@test}{\else \NAT@nm}{\else \NAT@nmfmt{\NAT@nm}}{}{}

\DeclareRobustCommand\citepos
  {\begingroup
   \let\NAT@nmfmt\NAT@posfmt
   \NAT@swafalse\let\NAT@ctype\z@\NAT@partrue
   \@ifstar{\NAT@fulltrue\NAT@citetp}{\NAT@fullfalse\NAT@citetp}}

\let\NAT@orig@nmfmt\NAT@nmfmt
\def\NAT@posfmt#1{\NAT@orig@nmfmt{#1's}}
\makeatother

\begin{document}

\maketitle

\begin{abstract}
Algorithmic fairness has received considerable attention due to the failures of various predictive AI systems that have been found to be unfairly biased against subgroups of the population. Many approaches have been proposed to mitigate such biases in predictive systems, however, they often struggle to provide accurate estimates and transparent correction mechanisms in the case where multiple sensitive variables, such as a combination of gender \emph{and} race, are involved. This paper introduces a new open source \proglang{Python} package, \pkg{EquiPy}, which provides a easy-to-use and model agnostic toolbox for efficiently achieving fairness across multiple sensitive variables. It also offers comprehensive graphic utilities to enable the user to interpret the influence of each sensitive variable within a global context. \pkg{EquiPy} makes use of theoretical results that allow the complexity arising from the use of multiple variables to be broken down into easier-to-solve sub-problems. We demonstrate the ease of use for both mitigation and interpretation on publicly available data derived from the US Census and provide sample code for its use.
\end{abstract}

\section[Introduction]{Introduction}\label{sec:intro}

With the increased usage of machine learning (ML) models across different industries, the discovery of unjust biases in predictive models has also proliferated. Multiple cases have been prominently reported in the media, where examples include sexist recruiting algorithms, facial recognition systems that perform poorly for females with darker skin and challenges in recognizing specific subgroups in self-driving cars~\citep{goodall2014ethical, nyholm2016ethics,dastin2022amazon,buolamwini2018gender}. Though none of these underlying models aimed to be biased in the first place, the predictions were found to replicate or even exaggerate existing biases in the training data. Simply not using a sensitive variable, such as gender or race, is not sufficient to avoid such biases, as more complex ML models can simply start to proxy for the omitted variables \citep{obermeyer2019}. This has shown the need to develop metrics and methods that can systematically measure and mitigate such biases in a more consistent manner. 

In recent years, many approaches to address this issue have been developed. Broadly speaking, these methods can be categorized into pre-processing, in-processing, and post-processing methods. Pre-processing ensures fairness in the input data by removing biases within the data \citep{park2021learning,qiang2022counterfactual,10.5555/3455716.3455958}, in-processing methods incorporate fairness constraints during model training \citep{Agarwal_Beygelzimer_Dubik_Langford_Wallach18, wang2020towards, joo2020gender}, where fairness constraints are usually incorporated into the loss function, and post-processing methods (of which \pkg{EquiPy} makes use), that achieve fairness through modifications of the final scores \citep{karako2018using, kim2019multiaccuracy, zeng2022}. Fairness of the resulting corrected predictions is then evaluated with respect to a \textit{group fairness metric}, such as independence, separation, or sufficiency \citep{barocas-hardt-narayanan}. Here, we focus on the independence criterion, which enforces Demographic Parity (DP) by requiring similar prediction distributions across sensitive groups.

Of particular importance for the development of this package is the literature using optimal transport, a mathematical framework for measuring distributional differences. Intuitively, achieving DP-fairness in predictions involves \emph{transporting} unfair scores to fair ones, while minimizing the effects of this intervention to maintain predictive accuracy. In regression, methods like \cite{Chzhen_Denis_Hebiri_Oneto_Pontil20Wasser} and \cite{gouic2020projection} minimize the Wasserstein distance across groups to reduce discrimination. Similarly, in classification, \cite{chiappa2020general} and \cite{gaucher2023fair} leverage optimal transport to achieve fair scores. Building upon this studies, \cite{hu2023fairness} achieved fairness in multi-task learning through a task-wise correction of the scores and extended the results to a sequential correction in \cite{hu2023sequentially}. \pkg{EquiPy} incorporates these insights and ensures DP-fairness in ML scores from binary classifiers or regression models, first for a single sensitive attribute using the Wasserstein barycenter-based mitigation of \cite{Chzhen_Denis_Hebiri_Oneto_Pontil20Wasser} and then provides an extension to multiple discrete sensitive attributes via a sequential correction.  
This sequential correction approach facilitates not only an efficient correction, but also provides interpretable pathways of the correction methods which allows for a more in-depth analysis of its impacts. \pkg{EquiPy}'s foundation in optimal transport theory provides a strong theoretical basis, guarantees of optimality, and a natural extension to related applications such as causal inference and interpretability \citep{charpentier2023optimal, ratz2023addressing}.


\subsection{Related Software}

Tackling unfairness in ML algorithms 
has been broadly studied and tested on real datasets in recent years. To facilitate this task in practice, multiple packages in different programming languages have been introduced for this task, for example \pkg{FairLens} in \proglang{Python}, the \pkg{fairness}~\citep{fairness:2021} and \pkg{fairadapt}~\citep{fairadapt} packages in \proglang{R} or the multi-platform package \pkg{AIF360}~\citep{bellamy2019ai}. More general in its applications is the \pkg{FairLearn} package~\citep{weerts2023fairlearn} in \proglang{Python}, which provides a broad introduction into the field of algorithmic fairness and a demonstration on how to use different metrics and mitigation techniques implemented within. Within \pkg{FairLearn}, the bulk of the mitigation techniques can be classified as either pre- or in-processing techniques\footnote{With the exception being the implementation of a post-processing method developed by~\cite{hardt2016equality}, which crucially also requires the label (or regressand) to achieve fairness---a restriction that is not required in \pkg{EquiPy}.} that focus on achieving fairness either before or during the training of a model. More recently, \pkg{OxonFair}, introduced by \cite{delaney2024oxonfair}, enables the evaluation and enforcement of fairness in binary classification through group-specific thresholds, with support for NLP and Computer Vision applications.

\textcolor{black}{\pkg{EquiPy}, being a post-processing optimal transport-based procedure, adjusts scores from any binary classification or regression model at a guaranteed minimal cost to predictive accuracy to achieve DP. Its lightweight, model-agnostic design ensures compatibility with models trained in languages beyond \proglang{Python}, making fairness considerations accessible even in industries reliant on commercial software for their predictions. With simple \code{fit} and \code{predict} methods, it lowers the barrier to entry for analysts and practitioners, aligning with recent high-level packages \cite{pappalardo2022scikit, boudt2022analyzing, tierney2023expanding}. The custom-build visualization module also allows for the analysis of various policy-driven fairness scenarios, allowing non-technical decision-makers to understand and evaluate different approaches to achieving fairness and equity in predictive analytics, while illustrating iterative fairness-performance trade-offs arising from corrections across multiple sensitive attributes. Moreover, unlike \pkg{FairLearn} and \pkg{OxonFair}, the main method calibrates using only model predictions and sensitive variable(s), requiring no access to labels, facilitating seamless deployment in existing pipelines. Finally, given that most of the implementation of \pkg{EquiPy} does not rely on purpose-built software nor specialised hardware, but simple high-level mathematical operations, a future implementation in other programming languages such as \proglang{R} should therefore not pose significant compatibility issues.}

A stable release can be installed via the \proglang{Python} Package index using
\begin{verbatim}
    pip install equipy
\end{verbatim}
The development version of the package is available on GitHub. A preliminary version can be installed directly from the GitHub repository using
\begin{verbatim}
    pip install --upgrade git+https://github.com/equilibration/equipy.git
\end{verbatim}

\paragraph*{Paper outline.} This paper is structured as follows: In Section~\ref{sec:AFOT}, we provide a brief introduction to algorithmic fairness and the optimal transport theory, establishing connections with the \pkg{EquiPy} package. Sections~\ref{sec:models} and \ref{sec:description} introduce respectively the main mathematical propositions and functionalities of \pkg{EquiPy}, mainly derived from \cite{hu2023sequentially}.
In Section~\ref{sec:illustrations}, we present examples of these functions through a case study on an open-source dataset. Finally, Section~\ref{sec:summary} summarizes our contribution and offers concluding remarks.

\section{Algorithmic Fairness and Optimal Transport}
\label{sec:AFOT}

Algorithmic fairness has close links to the field of optimal transport, to which we provide a short formal discussion below. As an informal introduction to the topic, we focus on the concept of \emph{Demographic Parity} (DP), also formally defined below, to establish fairness\footnote{DP is probably also the most commonly used fairness metric. Since \pkg{EquiPy} is intended for usage on both classification \emph{and} regression tasks, we restrict the analysis on this metric, as other common metrics such as Equalized Odds are not directly applicable to the regression case.}. Intuitively speaking, DP requires the predictions of a model to be independent of a sensitive variable. As other variables used within a predictive model often correlate with the sensitive one, independence of the scores goes beyond isolating the direct effects of a sensitive variable. That is, DP-fair scores should be independent of the direct effects as well as the indirect effects of the variable. As an example, body height is generally strongly correlated with sex, a DP-fair prediction on the sensitive variable \emph{sex} would therefore also need to correct for the discrepancies arising due to different distributions of height between the sexes. This also explains why it is not sufficient to simply omit the sensitive attribute, as a model can simply learn to proxy using a mix of related variables, to which ML models with complex feature interactions are especially prone to. 

\begin{figure}
    \centering
    \includegraphics[width=0.6\linewidth]{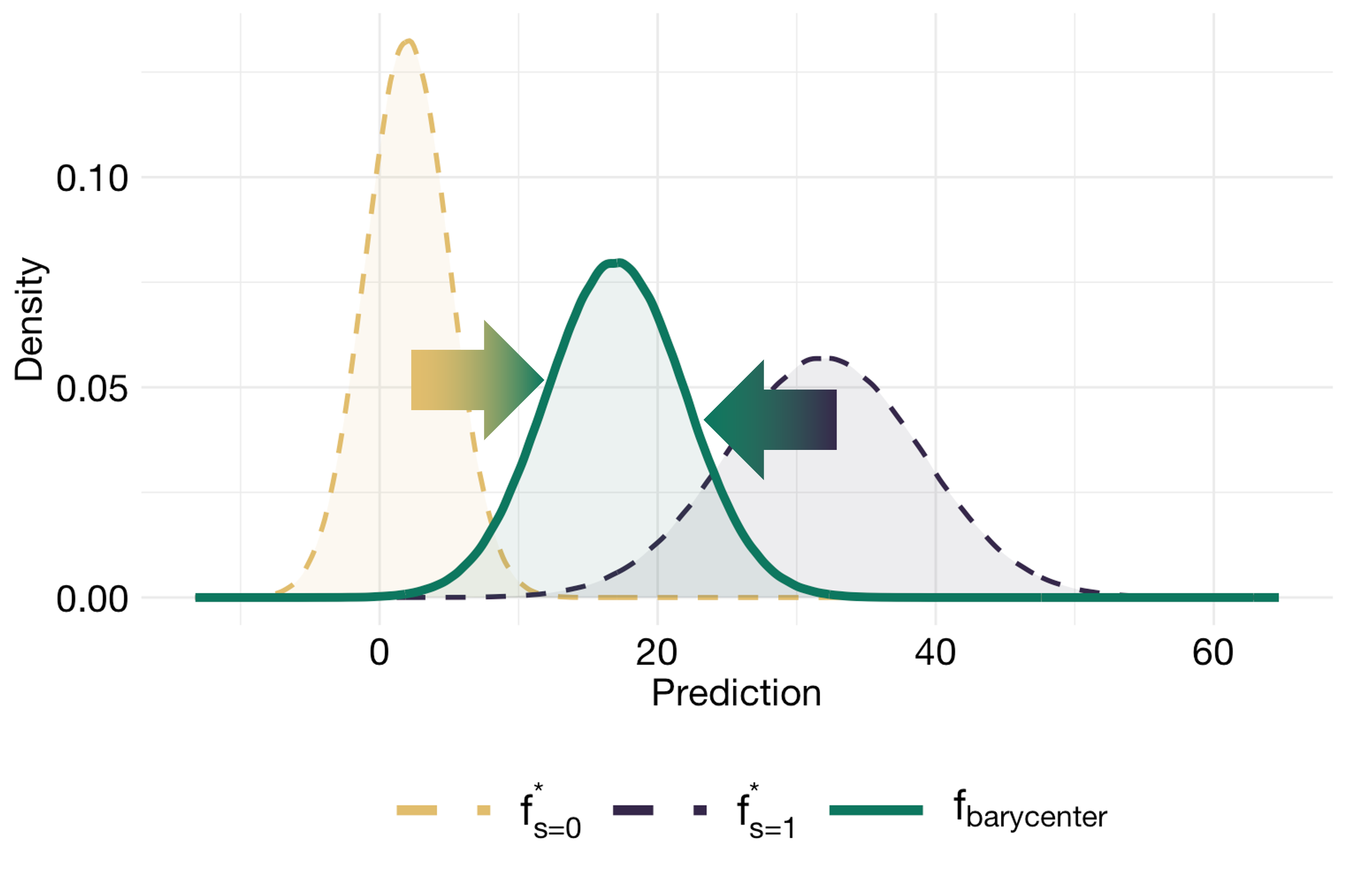}
    \caption{Group-wise distribution of predictions and barycenter. Note that the $y$-axis represents the density and the barycenter contains the both of the group wise predictions. }
    \label{fig:ex_barycenter}
\end{figure}

To see how transporting scores could help achieve DP, consider Figure \ref{fig:ex_barycenter}. Suppose that we have trained a model $f^*$ and analyze the distribution of the predictions given some sensitive variable $S\in\lbrace 0,1\rbrace$, resulting in the yellow and purple density curve. Clearly, DP would not be satisfied in this case. If instead the predictions were grouped in a single ``{middle-ground}'' distribution that builds a form of center for both group-wise prediction distributions, DP would be satisfied. This middle ground can be calculated using a Wasserstein barycenter, formally discussed below, and comes with a number of handy properties, such as being the distribution that satisfies DP but also minimizes the transportation cost. Intuitively, minimizing the transportation cost is a desirable property, as any change to the optimal predictions (that is, the initial \emph{unfair} predictions) will necessarily lead to a degradation in predictive metrics. By keeping the changes to the original predictions as small as possible, this negative impact is in turn minimized. 

\subsection{Background on Optimal Transport}

In this Section, we provide a brief formal introduction to the concept of the Wasserstein barycenter, emphasizing its characteristics within one-dimensional optimal transport theory and its correlation with algorithmic fairness. For those seeking further insights, excellent resources such as \cite{santambrogio2015optimal, villani2021topics} are available for an overview of optimal transport theory and \cite{chiappa2020general, Chzhen_Denis_Hebiri_Oneto_Pontil20Wasser} provide a more specific overview of its implications in algorithmic fairness.

Throughout this article, we refer to $\mathcal{V}$ as the space of univariate probability measures on $\mathcal{Y}\subset \mathbb{R}$ with finite variance (see \cite{agueh2011barycenters} for technical assumptions in general cases). Let $\nu_{1}$ and $\nu_{2}$ represent two probability measures in $\mathcal{V}$. A commonly employed method for measuring the difference between distributions involves the \textit{Wasserstein distance}, which computes the minimum ``cost'' required to transform one distribution into the other. In our case, we are interested in the $p$-Wasserstein distance (with $p\geq 1$) between $\nu_{1}$ and $\nu_{2}$, defined as:
\begin{equation*}
    \mathcal{W}_p^p(\nu_{1}, \nu_{2}) = \inf_{\pi\in\Pi(\nu_{1}, \nu_{2})} \mathbb{E}_{(Z_1, Z_2)\sim \pi}\left|Z_2-Z_1\right|^p,
\end{equation*}
where $\Pi(\nu_{1}, \nu_{2})$ is the set of distributions on $\mathcal{Y}\times\mathcal{Y}$ having $\nu_{1}$ and $\nu_{2}$ as marginals. In the univariate case, this expression can be rewritten as
$$
\mathcal{W}_p^p(\nu_{1}, \nu_{2}) = \int_{u\in [0, 1]} \left| Q_1(u) - Q_2(u) \right|^p du\enspace,
$$
where $Q_1$ and $Q_2$ are the associated quantile functions of $\nu_1$ and $\nu_2$ respectively. A coupling that achieves this infimum is called the optimal coupling between $\nu_{1}$ and $\nu_{2}$.
An alternative way to look at this problem is to seek for a mapping $T$, solution of
$$
\inf_{T:T_\#\nu_1=\nu_2} \mathbb{E}_{Z_1\sim \nu_1} |T(Z_1)-Z_1|^p
$$
where $T_\#\nu_1=\nu_2$ means that we consider push-forward measures of $\nu_1$ onto $\nu_2$. If $\nu_1$ is non-atomic and $p\geq 1$, there exists an optimal deterministic mapping $T^\star$. And if $\nu_1$ and $\nu_2$ are univariate measures in a subset of $\mathbb{R}$ (as considered here), and $p>1$, it is unique, and given by $T^\star(z_1)=Q_{|2}\circ F_{|1}(z_1)$ where $F_{|1}$ is the cumulative distribution function associated with $\nu_1$ and $Q_{|2}$ is the quantile function associated with $\nu_2$. If $p=1$, that mapping is not unique, but it is still optimal.

\paragraph*{Wasserstein Barycenter}
Throughout this article, we will frequently make use of \textit{Wasserstein Barycenters} \citep{agueh2011barycenters}. The Wasserstein Barycenter finds a representative distribution that lies between multiple given distributions in the Wasserstein space. It is defined for a family of $K$ measures $(\nu_1, \dots, \nu_K)$ in $\mathcal{V}$ and some positive weights $(w_1, \dots, w_K) \in\mathbb{R}_+^K$. The Wasserstein barycenter, denoted as ${\rm Bar}\{(w_k, \nu_k)_{k=1}^K\}$ is the minimizer
\begin{equation}\label{eq:WB}
    {\rm Bar}\{(w_k, \nu_k)_{k=1}^K\} = \argmin{\nu\in\mathcal{V}}\sum_{k=1}^{K}w_k\cdot \mathcal{W}_2^2\left( \nu_k, \nu\right)\enspace.
\end{equation}

The work in \cite{agueh2011barycenters} demonstrates that in our setup, the barycenter exists, and a sufficient condition for uniqueness is that one of the measures $\nu_i$ has a density with respect to the Lebesgue measure. For the sake of simplicity and readability, we assume that this condition holds for all distributions introduced in this article. It is worth noting that intuitively, comparing distributions $f(\boldsymbol{X}, \boldsymbol{A}) | \boldsymbol{A} = \boldsymbol{a}$ with $\boldsymbol{A}$ the sensitive attributes, where $\boldsymbol{a}\in\mathcal{A}$, and $f$ a given predictor seems natural. This concept becomes clearer in the subsequent sections, where we explore how the notion of unfairness is closely linked to these group-wise predictor response distributions. 


\paragraph*{Notation} Following the notation introduced in \cite{hu2023sequentially}, we examine a predictor $f:\mathcal{X}\times \mathcal{A}\to \mathcal{Y}$ where $(\boldsymbol{X}, \boldsymbol{A}, Y)$ is a random tuple drawn from the probability distribution $\mathbb{P}$. Here, $\boldsymbol{X} \in \mathcal{X} \subset \mathbb{R}^d$ represents the $d$ features, $\boldsymbol{A} = (A_1, A_2, \ldots, A_r) \in \mathcal{A} := \mathcal{A}_1\times\mathcal{A}_2\times\ldots\times \mathcal{A}_r\subset \mathbb{N}^r$ represents the $r$ sensitive features (or attributes), each of which is discrete and subject to fairness enforcement. For convenience, we define $A_{i:i+k} := (A_i, A_{i+1}, \cdots, A_{i+k})$ as the sequence of $k+1$ sensitive attributes ranging from $i$ to $i+k$. Thus, $\boldsymbol{A} = A_{1:r}$. The task to be estimated is represented by $Y \in \mathcal{Y}\subset \mathbb{R}$ and may involve either regression or a probability score for classification. Further, additional quantities are introduced below. Given $\boldsymbol{A}=(A_1, \ldots, A_i, \ldots, A_r)$ and $\boldsymbol{a} = (a_1, \ldots, a_i, \ldots, a_r) \in \mathcal{A}$, we denote them as follows:

\begin{itemize}
    \item $\nu_f$ the probability measure of $f(\boldsymbol{X}, \boldsymbol{A})$ in $\mathcal{V}$;

    \item $\nu_{f|\boldsymbol{a}}$ and $\nu_{f|a_i}$ (in $\mathcal{V}$) the probability measure of $f(\boldsymbol{X}, \boldsymbol{A}) | \boldsymbol{A} = \boldsymbol{a}$ and $f(\boldsymbol{X}, \boldsymbol{A}) | A_i = a_i$ respectively;

    \item $F_{f|\boldsymbol{a}}(u) := \mathbb{P}(f(\boldsymbol{X}, \boldsymbol{A})\leq u | \boldsymbol{A} = \boldsymbol{a})$ and, with an abuse of notation, $F_{f|a_i}(u) := \mathbb{P}(f(\boldsymbol{X}, \boldsymbol{A})\leq u | A_i = a_i)$ as their cumulative distribution function (CDF);

    \item $Q_{f|\boldsymbol{a}}(v) := \inf\{u\in\mathbb{R}: F_{f|\boldsymbol{a}}(u)\geq v\}$ and, with an abuse of notation, $Q_{f|a_i}(v) := \inf\{u\in\mathbb{R}: F_{f|a_i}(u)\geq v\} $  as their associated quantile functions.
\end{itemize}

\subsection{Definitions of Unfairness}\label{sec:unfairness}
As described earlier {Demographic Parity} is used to determine the fairness of a predictor, and is applicable to both classification and regression tasks. \pkg{EquiPy} is able to handle the presence of multiple sensitive attributes, and hence as a special case also the situation where a single sensitive attribute is present. Since most of the existing literature does not consider the more general case, we extend from the standard case to define unfairness in the context of multiple sensitive attributes. 

\paragraph*{Single sensitive attribute (SSA).}
\label{para:ssa}
In algorithmic fairness literature regarding a sensitive attribute $A_i$, the DP-unfairness requires $f(\boldsymbol{X}, \boldsymbol{A}) \indep A_i$ and its measure is typically defined with the following Total Variation,
$$
\mathcal{U}_{TV}(f) = \max_{a_i\in\mathcal{A}_i} \sup_{I\subset \mathbb{R}}\left| \mathbb{P}(f(\boldsymbol{X}, \boldsymbol{A}) \in I|A_i=a_i) - \mathbb{P}(f(\boldsymbol{X}, \boldsymbol{A}) \in I) \right|\enspace,
$$
or defined based on the Kolmogorov-Smirnov test, 
$$
\mathcal{U}_{KS}(f) = \max_{a_i\in\mathcal{A}_i}\sup_{u\in \mathbb{R}} \left| F_{f}(u) - F_{f|a_i}(u) \right| \enspace.
$$
In our study, the unfairness measure of the predictor on the feature $A_i$ is given by the corresponding 1-Wasserstein based measure:
\begin{equation}\label{eq:UnfairnessSingle}
\mathcal{U}_i(f) = \max_{a_i\in \mathcal{A}_i} \mathcal{W}_1(\nu_f, \nu_{f|a_i}) = \max_{a_i\in \mathcal{A}_i}\int_{u\in[0,1]} \left|\ Q_{f}(u) - Q_{f|a_i}(u)\ \right|du\enspace.
\end{equation}

We say that the predictor $f$ is fair with respect to $A_i$ under Demographic Parity if and only if $\mathcal{U}_{\square} (f) = 0$, where $\square\in\{TV, KS, i\}$. This also ties in with the Wasserstein Distance, as it is clear that the DP condition is satisfied if and only if the Wasserstein distance between the group-wise distributions is zero. To facilitate both the attribution of each individual sensitive variable to a global measure of unfairness of a predictor, we opt for an additive measure of unfairness.

\paragraph*{Multiple sensitive attributes (MSA).}
\label{para:msa}
For the multiple sensitive attributes $A_{i}, \dots, A_{i+k}$, their collective unfairness is assessed simply by summing the unfairness of each marginal sensitive attribute:
\begin{equation}\label{eq:UnfairnessMulti}
    \mathcal{U}_{\{i, \dots, i+k\}}(f) = \mathcal{U}_{i:i+k}(f) = \mathcal{U}_i(f) + \dots + \mathcal{U}_{i+k}(f)\enspace.
\end{equation}
Note here that from the definition of the unfairness measure in the MSA context, an ordered sequence of fairness variables is supposed. This will be important later on, as it allows to \emph{sequentially} render fair a model. 

Within \pkg{EquiPy} the measure for both SSA and MSA is implemented in the function\\\code{unfairness}, which expects a vector of scores of size $N \times 1$ (\code{predictions} in the example below) and a matrix of size $N\times r$, with $N$ the number of observations and $r$ the number of sensitive attributes (\code{sensitive_features} in the example), that can be either in the form of a dataframe \pkg{pandas} or an array using \pkg{numpy}. We recommend using \pkg{pandas} objects to specify the sensitive attribute names, facilitating graph reading.


\begin{verbatim}
    import numpy as np
    import pandas as pd
    from equipy.metrics import unfairness

    predictions = np.array([0.05, 0.08, 0.9, 0.5, 0.18, 0.92, 0.9, 0.5, 
        0.16, 0.79])
    sensitive_features = pd.DataFrame({'origin': [1, 0, 0, 1, 1, 1, 0, 0, 
        0, 1], 'gender': [1, 1, 1, 0, 0, 1, 0, 0, 0, 1]})
    unfairness(predictions, sensitive_features)
    >>> 0.472
\end{verbatim}

Under the hood, \pkg{EquiPy} can calculate the unfairness metric directly using the optimal transportation plan through the option \code{approximate=False}, but as this requires solving an $N \times N$ linear program, the default method approximates the transportation cost using a grid on the empirical quantile functions specified in Equation \eqref{eq:UnfairnessSingle}, which is computationally less demanding.

\subsection{Optimal Fair Projection} \label{optimal-fair-projection}

The core problem within algorithmic fairness is to conduct the trade-off between model accuracy and minimizing unfairness. Having a constant predictor would ensure fairness across the predictions but is certainly undesirable from the view of predictive accuracy. To address this issue, the literature has developed tools to obtain optimal predictive accuracy under the fairness constraint. To both quantify and mitigate predictions from a model, we consider $f^*(\boldsymbol{X}, \boldsymbol{A}):= \mathbb{E}[Y|\boldsymbol{X}, \boldsymbol{A}]$ the Bayes rule that minimizes the squared risk 
\begin{equation}
\label{performance}
  \mathcal{R}(f):=\mathbb{E}(Y-f(\boldsymbol{X}, \boldsymbol{A}))^2\enspace, 
\end{equation}
 as the \emph{optimal} model,
 $$
f^\star:=\underset{f\in\mathcal{F}}{\text{argmin}}\big\{{\mathcal{R}}(f)\big\} 
 $$ 
In practical terms, this represents the model estimated on the data without any constraints. To achieve fair predictions, restrictions need to be imposed on the class of available models, denoted $\mathcal{F}$. 

\paragraph{SSA case.}
We define the class of DP-fair models for the $i$-th sensitive feature $A_i$,
$$
\mathcal{F}_{\mathrm{fair}, i} := \big\lbrace f\in\mathcal{F}\text{ s.t. }f(\boldsymbol{X}, \boldsymbol{A})\indep A_i \big\rbrace = \big\lbrace f\in\mathcal{F}\text{ s.t. }\mathcal{U}_i(f) = 0 \big\rbrace \subset \mathcal{F} \enspace.
$$
Fairness for the feature $A_i$ is achieved by projection onto a fair subspace, resulting in a fair predictor $f_{B_i}$
$$
f_{B_i} \in \underset{f\in\mathcal{F}_{\mathrm{fair}, i}}{\text{argmin}}\big\{\mathcal{R}(f)\big\}\enspace. 
$$ 
Given a risk $\mathcal{R}$, a class $\mathcal{F}$ and the fair subclass $\mathcal{F}_{\mathrm{fair}, i}$ for $A_i$, the price of fairness is quantified by
$$
\mathcal{E}_{\mathrm{fair}, i}(\mathcal{F}) = \underset{f\in\mathcal{F}_{\mathrm{fair}, i}}{\text{min}}\big\{{\mathcal{R}}(f)\big\} -
\underset{f\in\mathcal{F}}{\text{min}}\big\{{\mathcal{R}}(f)\big\} = \underset{f\in\mathcal{F}_{\mathrm{fair}, i}}{\text{min}}\big\{{\mathcal{R}}(f)\big\} -\mathcal{R}(f^*)\enspace.
$$
\cite{Chzhen_Denis_Hebiri_Oneto_Pontil20Wasser} and \cite{gouic2020projection} have both demonstrated that in the SSA case, under the DP notion of fairness, the price of fairness for $A_i$ corresponds to the following Wasserstein barycenter:
\begin{equation}\label{eq:ExcessRiskParam}
    \mathcal{E}_{\mathrm{fair}, i}(\mathcal{F}) = \inf_{f \in \mathcal{F}_{\mathrm{fair}, i}}\sum_{a_i\in\mathcal{A}_i}p_{a_i}\cdot \mathcal{W}_2^2\left(\nu_{f^*|a_i}, \nu_{f}\right)\enspace,
\end{equation}
where $p_{a_i} := \mathbb{P}(A_i=a_i)$, and its minimum is achieved when using (see Equation~\eqref{eq:WB})\\${\rm Bar}( p_{a_i}, \nu_{f^*|a_i})_{a_i\in\mathcal{A}_i}$---also denoted as $\mu_{\mathcal{A}_i}(\nu_{f^*})$, with $\nu_{f^*}$ representing the measure associated with the optimal (DP-unconstrained) predictor $f^*$.

\paragraph{MSA case.}

\cite{hu2023sequentially} extends these formulations to the case of multiple sensitive attributes, where if we denote $\mathcal{F}_{\text{fair}} := \big\lbrace f\in\mathcal{F}~|~\mathcal{U}(f) = 0 \big\rbrace \subset \mathcal{F}$ as the class of DP-fair models for all sensitive attributes $\boldsymbol{A} = (A_1, \ldots, A_r)$ (or a subset), we have in summary the corresponding:
$$
\begin{cases}
\text{fair predictor:}\quad f_{B} \in \underset{f\in\mathcal{F}_{\mathrm{fair}}}{\text{argmin}}\big\{\mathcal{R}(f)\big\}\\
\text{price of fairness:}\quad \mathcal{E}_{\mathrm{fair}}(\mathcal{F}) := \underset{f\in\mathcal{F}_{\mathrm{fair}}}{\text{min}}\big\{{\mathcal{R}}(f)\big\}-\underset{f\in\mathcal{F}}{\text{min}}\big\{{\mathcal{R}}(f)\big\} =\mathcal{R}(f_B)-\mathcal{R}(f^*)\\
\text{optimal fair distribution:}\quad \mu_{\mathcal{A}}(\nu_{f^*}) := {\rm Bar}( p_{\boldsymbol{a}}, \nu_{f^*|\boldsymbol{a}})_{\boldsymbol{a}\in\mathcal{A}}
\enspace.
\end{cases}
$$
In this context, $p_{\boldsymbol{a}} := \mathbb{P}(\boldsymbol{A}=\boldsymbol{a})$. Throughout this article, we consistently refer to $f_{B_i}$ as the aforementioned fair predictor in SSA and $f_{B}$ as the above overall fair predictor, both derived through the Wasserstein barycenter method. The associated measures are denoted respectively as $\nu_{f_{B_i}} = \mu_{\mathcal{A}_i}(\nu_{f^*})$ and $\nu_{f_B} = \mu_{\mathcal{A}}(\nu_{f^*})$.


Drawing upon \cite{Chzhen_Denis_Hebiri_Oneto_Pontil20Wasser} for SSA and \cite{hu2023sequentially} for MSA, the fair predictor \( f_B \) is considered optimal as it minimizes the risk \( \mathcal{R} \) among fair predictors. Its closed-form solution can be expressed as:
\begin{equation}\label{eq:OptFairGlobal}
    f_{B}(\boldsymbol{x}, \boldsymbol{a}) = \left( \sum_{\boldsymbol{a}'\in\mathcal{A}} p_{\boldsymbol{a'}}
    Q_{f^*|\boldsymbol{a}'}\right)\circ F_{f^*|\boldsymbol{a}}\left( f^*(\boldsymbol{x}, \boldsymbol{a}) \right) \quad \text{for all }(\boldsymbol{x}, \boldsymbol{a})\in\mathcal{X}\times \mathcal{A}\enspace.
\end{equation}

\paragraph{In practice.} Within the context of SSA, DP-fairness can be achieved using a straightforward procedure. \pkg{EquiPy} offers the class \code{FairWasserstein} to calibrate DP-fairness by internally calculating the empirical counterparts of $p_{\boldsymbol{a}}$, $F_{f^*|\boldsymbol{a}}$, and $Q_{f^*|\boldsymbol{a}}$ in Equation \eqref{eq:OptFairGlobal}, where $\boldsymbol{A} \in \mathcal{A}$ contains a unique sensitive attribute here. The class expects as input the sensitive variable as a vector, represented by \code{sensitive_feature_calib} either using \pkg{pandas} or \pkg{numpy}, and the predictions \code{predictions_calib} of an estimator of $f^*$. Again, the estimator can be any trained ML algorithm. DP-fair predictions can then be obtained by transforming given predictions from a test-set. The resulting empirical formulation is implemented in the \pkg{EquiPy} package and can be accessed as follows:

\begin{verbatim}
    # Single Sensitive Attribute (SSA) case
    import numpy as np
    import pandas as pd
    from equipy.fairness import FairWasserstein

    calibrator = FairWasserstein(sigma = 0.0001)

    # calibration set
    predictions_calib = np.array([0.05, 0.08, 0.9, 0.5, 0.18, 0.92, 0.9, 
        0.5])
    sensitive_feature_calib = pd.DataFrame({'origin': [1, 0, 0, 1, 1, 1, 
        0, 0]})
    calibrator.fit(predictions_calib, sensitive_feature_calib)

    # studied set
    predictions = np.array([0.16, 0.79])
    sensitive_feature = pd.DataFrame({'origin': [0, 1]})

    calibrator.transform(predictions, sensitive_feature)
    >>> array([0.271, 0.752])
\end{verbatim}

Note that setting \code{sigma = 0.0001}, which is the default value for \code{sigma}, corresponds to what is commonly referred to as the \emph{jittering} process. This involves introducing a small perturbation into a given predictor by adding continuous random noise. The purpose of this perturbation is to prevent the occurrence of \emph{atoms} in the predictions; for further details on regression, refer to \citep{Chzhen_Denis_Hebiri_Oneto_Pontil20Wasser}, and for classification, refer to \citep{denis2021fairness}. 



\section{Sequential Fairness and Software} \label{sec:models}

Whereas the application of the Wasserstein Barycenter is straightforward in the SSA context, multiple sensitive attributes can pose problems within both the estimation and interpretation of the effects of the projection. A simple possibility would be to recode multiple sensitive features into a single, multi-class sensitive feature. However, this introduces problems in the robustness of the estimation as it introduces exponentially many subgroups on a fixed size data set. Further, such a formulation hampers approximate fairness, discussed below. In line with the article by \cite{hu2023sequentially}, we present in this section a concise exposition of Equation~\eqref{eq:OptFairGlobal} to improve transparency regarding the inter-correlations among sensitive features. This breakdown facilitates a clearer understanding. Following this, we show how the \pkg{EquiPy} Package addresses these aspects in practical applications.

\subsection{A Sequentially Fair Mechanism}

Recall that $f_{B_i}$ represents a predictor that ensures fairness with respect to the sensitive attribute $A_i$. To simplify notation, we express the composition $f_{B_i} \circ f_{B_j}$ by a slight abuse of notation as:
\begin{equation}
    \left( f_{B_i}\circ f_{B_j} \right)(\boldsymbol{x}, \boldsymbol{a}) = \left( \sum_{a_i'\in\mathcal{A}_i} p_{a_i'}Q_{f_{B_j}|a_i'}\right)\circ F_{f_{B_j}|a_i}\left( f_{B_j|a_i}(\boldsymbol{x}, \boldsymbol{a}) \right) \quad \text{for all }(\boldsymbol{x}, \boldsymbol{a})\in\mathcal{X}\times \mathcal{A}\enspace.
    \label{eq:compo}
\end{equation}

Drawing on insights from the article by \cite{hu2023sequentially}, the overarching barycentric fair predictor $f_B$ within the space $\mathcal{A}$ can be articulated as a sequence of compositions of marginal ``barycentric'' fair predictors, advancing sequentially through $\mathcal{A}_1, \ldots, \mathcal{A}_r$. 
\begin{align*}
f_{B} (\boldsymbol{x}, \boldsymbol{a}) &= \left(f_{B_1}\circ f_{B_2}\circ \ldots \circ f_{B_r}\right)(\boldsymbol{x},\boldsymbol{a}) \quad \text{for all }(\boldsymbol{x}, \boldsymbol{a})\in\mathcal{X}\times \mathcal{A}\enspace,
\end{align*}
corresponding to the composition of measures: 
$\mu_{\mathcal{A}}(\nu_{f^*}) = \mu_{\mathcal{A}_1}\circ\mu_{\mathcal{A}_2}\circ\cdots\circ \mu_{\mathcal{A}_r}(\nu_{f^*})$.
It's important to note that the fairness mitigation remains unaffected by the order in which the barycentric mitigation is executed. Hence, for $r=2$ sensitive features, $f_{B_1}\circ f_{B_2} = f_{B_2}\circ f_{B_1}$. The proposition and its corresponding proof can be found in \citep{hu2023sequentially}.

\paragraph{In practice.} The sequential approach for multiple sensitive variables is implemented in the \code{MultiWasserstein} class, which internalizes the computations, enabling a straightforward adaption similar to the \code{FairWasserstein} class. It can be executed as follows:
\begin{verbatim}
    import numpy as np
    import pandas as pd
    from equipy.fairness import MultiWasserstein

    calibrator = MultiWasserstein(sigma = 0.0001)

    # calibration set
    predictions_calib = np.array([0.05, 0.08, 0.9, 0.5, 0.18, 0.92, 0.9, 
        0.5])
    sensitive_features_calib = pd.DataFrame({'origin': [1, 0, 0, 1, 1, 1, 
        0, 0], 'gender': [1, 1, 1, 0, 0, 1, 0, 0]})
    calibrator.fit(predictions_calib, sensitive_features_calib)

    # studied set
    predictions = np.array([0.16, 0.79])
    sensitive_features = pd.DataFrame({'origin': [0, 1], 
        'gender': [0, 1]})

    calibrator.transform(predictions, sensitive_features)

    >>> array([0.203, 0.361])
\end{verbatim}

For further details, we refer to the paper~\citep{hu2023sequentially} or the dedicated package documentation in \url{https://equilibration.github.io/equipy/}.

\subsection{Extensions to Approximate Fairness}
\label{subsec:approxfairness}

In our context, achieving \emph{approximate fairness} involves improving fairness relatively and approximately with a preset level of relative fairness improvement, denoted as $\boldsymbol{\varepsilon} = \varepsilon_{1:r} = (\varepsilon_{1}, \cdots, \varepsilon_{r})$ (abbreviated as $\boldsymbol{\varepsilon}$-fairness in this article). The SSA framework discussed in the article by \cite{chzhen2022minimax} employs geodesic parameterization to generate predictors that approximate fairness. Specifically, for any $A_i\in\mathcal{A}_i$ without loss of generality, the predictor achieving $\varepsilon_i$-fairness takes the form:
$$
f_{B_i}^{\varepsilon_i}(\boldsymbol{X}, \boldsymbol{A}) := (1-\varepsilon_i)\cdot f_{B_i}(\boldsymbol{X}, \boldsymbol{A}) + \varepsilon_i \cdot f^*(\boldsymbol{X}, \boldsymbol{A})\enspace.
$$
This predictor achieves (optimally) the following risk-fairness trade-off:
$$
f_{B_i}^{\varepsilon_i} \in \argmin{f\in\mathcal{F}}\{ \mathcal{R}(f): \mathcal{U}_i(f)\leq\varepsilon_i \cdot \mathcal{U}_i(f^*)\}\enspace.
$$
Quoting \citep{hu2023sequentially}, ``introducing a sequential approach is pivotal for enhancing clarity''. This introduced sequentially fair framework aids in comprehending intricate concepts such as the aforementioned approximate fairness within the MSA context. Indeed, as described in the article by \cite{hu2023sequentially}, if we permit

$$
f_{B}^{\boldsymbol{\varepsilon}} \in \argmin{f\in\mathcal{F}}\left\{ \mathcal{R}(f): \mathcal{U}(f) \leq \textstyle \sum_{i=1, \ldots, r}\varepsilon_i \cdot \mathcal{U}_i(f^*) \right\}\enspace,
$$
then $
f_{B}^{\boldsymbol{\varepsilon}} = f_{B_1}^{\varepsilon_1}\circ\,\cdots\,\circ f_{B_r}^{\varepsilon_r}$. This allows us to break down how various components of the sequential fairness mechanism interact to achieve fairness goals. It also facilitates the interpretation of the inherent effects of fairness adjustments.

\paragraph{In practice:} In \pkg{EquiPy}, the empirical counterpart of approximate fairness is inherently incorporated within both the \code{FairWasserstein} and \code{MultiWasserstein} classes of the \code{fairness} module. An illustrative example of its application is provided below:

\begin{verbatim}
    import numpy as np
    import pandas as pd
    from equipy.fairness import MultiWasserstein

    calibrator = MultiWasserstein(sigma = 0.0001)

    # calibration set
    predictions_calib = np.array([0.05, 0.08, 0.9, 0.5, 0.18, 0.92, 0.9, 
        0.5])
    sensitive_features_calib = pd.DataFrame({'origin': [1, 0, 0, 1, 1, 1, 
        0, 0], 'gender': [1, 1, 1, 0, 0, 1, 0, 0]})
    calibrator.fit(predictions_calib, sensitive_features_calib)

    # studied set
    predictions = np.array([0.16, 0.79])
    sensitive_features = pd.DataFrame({'origin': [0, 1], 
        'gender': [0, 1]})

    # approximate fairness
    epsilon = [0.1, 0.2]

    calibrator.transform(predictions, sensitive_features, 
        epsilon = epsilon)

    >>> array([0.210, 0.348])
\end{verbatim}

For additional details, we refer to the section below or the dedicated package documentation: \url{https://equilibration.github.io/equipy/}.

\begin{figure}
\begin{center}
\begin{forest}
  for tree={
    font=\scriptsize\ttfamily, 
    grow'=0,
    child anchor=west,
    parent anchor=south,
    anchor=west,
    calign=first,
    edge path={
      \noexpand\path [draw, line width=1.2pt, \forestoption{edge}] 
      (!u.south west) +(8pt,0) |- node[fill,inner sep=1pt] {} (.child anchor)\forestoption{edge label};
    },
    before typesetting nodes={
      if n=1
        {insert before={[,phantom]}}
        {}
    },
    fit=band,
    before computing xy={l=25pt}, 
    s sep=1pt, 
    level distance=6pt,
  }
[equipy
  [fairness
    [FairWasserstein
        [fit]
        [transform]
    ]
    [MultiWasserstein
        [fit]
        [transform]
    ]
  ]
  [metrics
    [unfairness]
    [performance]
  ]
  [graphs
    [fair\_arrow\_plot]
    [fair\_multiple\_arrow\_plot]
    [fair\_density\_plot]
    [fair\_waterfall\_plot]
  ]
]
\end{forest}
\caption{Tree structure of the \pkg{EquiPy} package}
\label{fig:treestructure}
\end{center}
\end{figure}
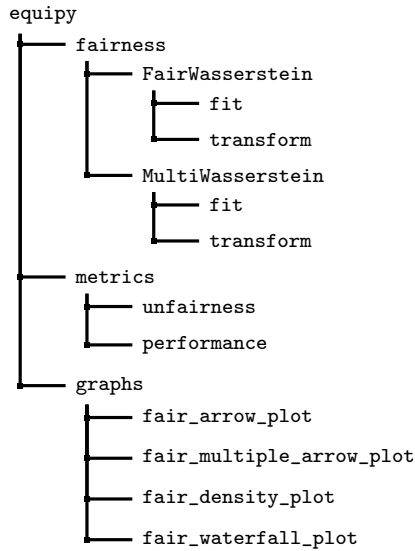

\section{Description of the EquiPy Package Structure}\label{sec:description}

The preceding sections offered an overview of the main modules and their functionalities illustrating the theoretical results. In this section, we will formally introduce the aforementioned \proglang{Python} classes and modules, along with all associated parameters (from version 0.0.11), before diving into practical illustrations demonstrating explicit applications and visual displays in Section~\ref{sec:illustrations}.


\subsection{Overview of the Package Structure}

The package is structured around three core modules: \code{fairness}, \code{metrics} and \code{graphs}, where each module contains several sub-modules that help the user access the relevant functions. Refer to Figure \ref{fig:treestructure} for an overview of the tree-structure. Additionally, Table \ref{tab:availablemethods} describes every sub-module in detail. Note that the the classes \code{FairWasserstein} and \code{MultiWasserstein} both contain a \code{fit} and \code{transform} method, similar to other packages used in ML, like \pkg{scikit-learn} in \proglang{Python}.

\begin{table}[H]
    \centering
    \begin{tabular}{lp{9.5cm}}
    \hline
    \textbf{Method} & \textbf{Description} \\ \hline
    \code{fit} & Fit distribution and quantile functions using the calibration data. Note that in practice, this calibration data could either be the training data or unlabeled data. \\
    \code{transform} & Transform data to enforce fairness with Wasserstein distance using results of \code{fit}. In practical applications, we typically apply it to a hold-out (or test) data.\\ \hline
    \code{unfairness} & Compute the unfairness measure (empirical version of Equation \ref{eq:UnfairnessMulti}) for a given fair output and multiple sensitive attributes data. \\
    \code{performance} & Compute the performance metric for predicted fair outputs compared to the true targets. \\ \hline
    \code{fair_arrow_plot} & Create an arrow plot illustrating the progression of \code{fairness}-\code{performance} combinations of a predictive model, step by step according to the sensitive attributes. \\
    \code{fair_multiple_arrow_plot} & This method uses \code{fair_arrow_plot} with respect to different permutations of sensitive attributes. \\
    \code{fair_density_plot} & Visualize the distribution of predictions conditional on different sensitive features using kernel density estimates. We use a Beta kernel when the predictive task is binary classification. \\
    \code{fair_waterfall_plot} & Generate a waterfall plot to visually represent how sequential fairness impacts a fairness metric associated with one or more sensitive attributes within a model.
    \\
    \hline
    \end{tabular}
    \caption{Overview of the core methods for the modules \code{fairness}, \code{metrics} and \code{graphs}}
    \label{tab:availablemethods}
\end{table}


\subsection{Description of the Package Modules and Sub-Modules}

The main modules follow the standard workflow in fairness mitigation. Mitigating biases through the \code{fairness} module, measuring both the predictive performance and unfairness using \code{metric} to enable a quantitative impact assessment and plotting utilities in \code{graphs}, for further interpretation and reporting. This subsection provides details on the three main modules, including comprehensive information about their parameters and validation functions.

\paragraph{Module fairness}

This module comprises two primary classes : \code{FairWasserstein}, ensuring fair predictions regarding a single sensitive attribute (Section~\ref{para:ssa}), and \code{MultiWasserstein}, with a similar structure but addressing fairness with respect to multiple sensitive attributes (Section~\ref{para:msa}). Both classes implement fairness adjustment on model predictions related to one or multiple sensitive attributes, using Wasserstein distance for binary classification and regression tasks, as introduced in Section \ref{sec:models}. In the case of binary classification, this class supports scores instead of predicted labels. A specific instance demonstrating the application of \code{FairWasserstein} and \code{MultiWasserstein} are presented respectively in Section~\ref{optimal-fair-projection} and Section~\ref{sec:models}. In addition, Figure \ref{fig:schemeWasserstein} illustrates the application process of \code{FairWasserstein} and \code{MultiWasserstein}, highlighting their model-agnostic nature. Specifically, the data not used during the training phase of the ML model is divided into calibration and test sets. The \code{fit} method is trained on the calibration dataset to learn the mathematical quantities from Equation \ref{eq:compo}, after which the \code{transform} method is applied to the test set  to obtain fair predictors using the functions learned during the calibration step. The resultant fair predictions, denoted as $\hat{y}^{\text{fair}}_{\text{test}}$, can subsequently be employed to calculate various methods within the \code{metrics} module. To enable approximate fairness, as described in Section ~\ref{subsec:approxfairness}, the user can specify values for the \code{epsilon} parameter, a vector of size $r$, in both \code{transform} functions of \code{FairWasserstein} and \code{MultiWasserstein}.



\tikzstyle{start} = [rectangle,
minimum width=4.8cm, 
minimum height=1cm,
text centered, 
draw=black, 
fill=white]

\tikzstyle{train} = [rectangle,
minimum width=4.8cm, 
minimum height=1cm,
text centered, 
draw=black, 
fill=white]

\tikzstyle{calib} = [rectangle,
minimum width=4.8cm, 
minimum height=1cm,
text centered, 
draw=black, 
fill=white]

\tikzstyle{test} = [rectangle,
minimum width=4.8cm, 
minimum height=1cm,
text centered, 
draw=black, 
fill=white]

\tikzstyle{fitml} = [rectangle,
minimum width=1.9cm, 
minimum height=0.5cm,
text centered, 
draw=white, 
fill=white]

\tikzstyle{calibscores} = [rectangle,
minimum width=1.9cm, 
minimum height=0.5cm,
text centered, 
draw=white, 
fill=white]

\tikzstyle{testscores} = [rectangle,
minimum width=1.9cm, 
minimum height=0.5cm,
text centered, 
draw=white, 
fill=white]

\tikzstyle{arrow} = [thick,->,>=stealth]

\tikzstyle{fit} = [rectangle, rounded corners, 
minimum width=2.5cm, 
minimum height=1cm,
text centered, 
draw=black, 
fill=white]

\tikzstyle{transform} = [rectangle, rounded corners, 
minimum width=2.5cm, 
minimum height=1cm,
text centered, 
draw=black, 
fill=white]

\tikzstyle{fairtest} = [rectangle,
minimum width=2cm, 
minimum height=0.5cm,
text centered, 
draw=white, 
fill=white]

\begin{center}
\begin{tikzpicture}[node distance=2cm]

\draw [draw=black, fill=white] (-3, -5) rectangle (12.5, 1);
\draw [draw=black, fill=white] (-3, -7) rectangle (12.5, -3);

\node (start) [start] {Dataset: $(\mathbf{x}, \mathbf{a}, y)$};
\node (train) [train, below of=start] {Train: $(\mathbf{x}, \mathbf{a},  y)_{\text{train}}$};
\node (fitml) [fitml, right of=train, xshift=2.5cm] {{$\hat{f}(\mathbf{x}, \mathbf{a})$}};
\node (calibscores) [calibscores, below of=fitml] {$\hat{y}_{\text{calib}}$};
\node (testscores) [testscores, below of=calibscores] {$\hat{y}_{\text{test}}$};
\node (fit) [fit, right of=calibscores, xshift=1.4cm] {\code{fit}};
\node (transform) [transform, below of=fit] {\code{transform}};
\node (calib) [calib, below of=train] {Calibration: $(\mathbf{x}, \mathbf{a},  y)_{\text{calib}}$};
\node (test) [test, below of=calib] {Test: $(\mathbf{x}, \mathbf{a},  y)_{\text{test}}$};
\node (fairtest) [fairtest, right of=transform, xshift=1.4cm] {$\hat{y}_{\text{test}}^{\text{fair}}$};

\node[above right] at (8.2, 0.3) {\textbf{Training ML model}};
\node[above right] at (10.8, -3.7) {\pkg{EquiPy}};

\draw [arrow, dashed] (calibscores) -- (testscores);
\draw [arrow] (start) -- (train) node[pos=0.5, anchor=west, inner sep=2pt, font=\small] {\textit{Split}};
\draw [arrow] (fitml) -- (calibscores) node[pos=0.3, anchor=west, inner sep=2pt, font=\small] {\textit{Predict}};
\draw [arrow] (fit) -- (calibscores) node[pos=0.5, above, font=\small] {\textit{Train}};
\draw [arrow] (fitml) -- (train) node[pos=0.5, above, font=\small] {\textit{Train}};
\draw [arrow] (transform) -- (testscores) node[pos=0.5, above, font=\small] {\textit{Apply}};
\draw [arrow] (fit) -- (transform) node[pos=0.5, anchor=west, inner sep=2pt, font=\small] {$\hat{f}_B(\mathbf{x}, \mathbf{a})$};
\draw [arrow] (transform) -- (fairtest);
\draw [dashed] (calib) -- (calibscores);
\draw [dashed] (test) -- (testscores);
\draw [arrow, dashed] (train) -- (calib);
\draw [arrow, dashed] (calib) -- (test);
\end{tikzpicture}

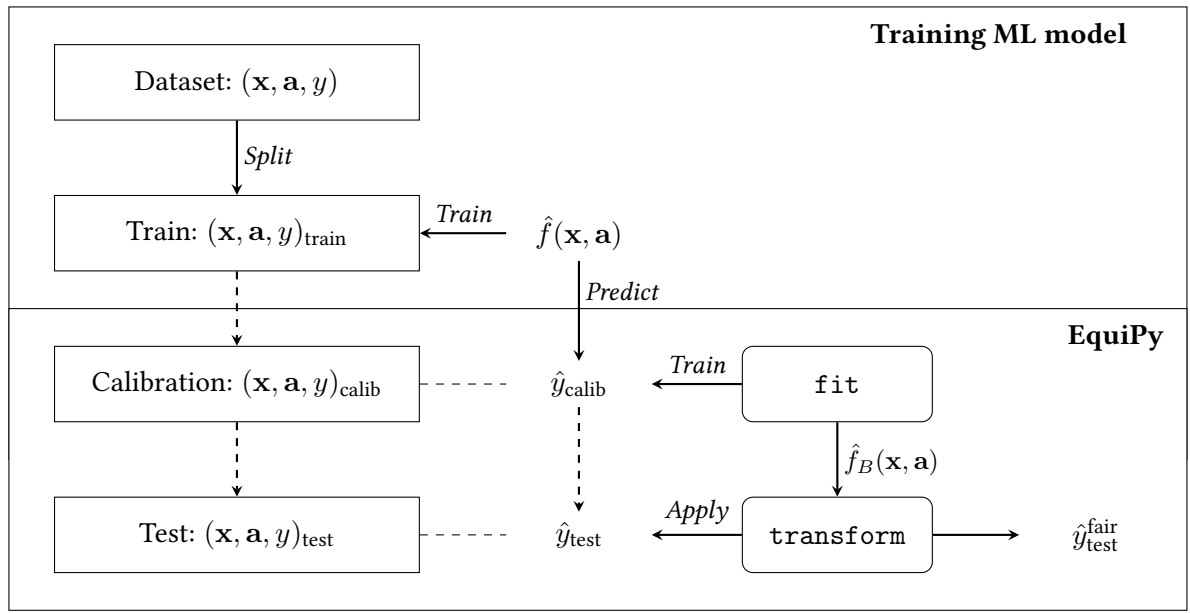
\captionof{figure}{Process of mitigating predictions using the methods of \code{MultiWasserstein} class}
\label{fig:schemeWasserstein}
\end{center}

\paragraph{Module metrics}

The \pkg{EquiPy} package provides functionality for computing two types of metrics : performance and unfairness. Performance metrics, calculated through the method \code{performance}, refer to commonly used regression or classification metrics such as\\\code{accuracy_score} or \code{mean_squared_error} from \code{sklearn.metrics} module but also allow custom user specification. The \code{performance} method provides a wrapper around the specified functions which allows a seamless transition between the \code{metrics} and \code{graphs} modules. It expects as inputs predictions and labels in array form, compatible with the underlying performance score. Hence for classification metrics, the predicted scores need to be transformed to label predictions first. The unfairness metric, as defined by Equation \eqref{eq:UnfairnessMulti} in Section \ref{sec:AFOT}, can be computed using the function \code{unfairness}. Here, the sensitive variables need to be provided as a \code{pandas.DataFrame} or a \code{numpy.array} object. The unfairness metric calculation is either computed using the Wasserstein distance from the \pkg{POT} package or either determined as the maximum difference in quantiles defined on a grid on [0,1] between the two populations. The default approach is the latter (\code{approximate=True}), and we recommend using the \pkg{POT} package (setting \code{approximate=False} in the \code{unfairness} function) when the number of observations is  insufficient to efficiently approximate the quantile functions by sensitive group.
A practical example of the usage is provided in Section \ref{sec:unfairness}. As an extension, the module also includes hidden methods used in the \code{graphs} module to provide an overview and interpretations of variations in unfairness and performance across different permutations of the sensitive variables. 

\paragraph{Module graphs} This module provides utilities for the visualization of the correction paths using three different graph types. The method \code{fair_arrow_plot} illustrates the fairness-performance relationship based on the ordered sequence of sensitive attributes specified by the user. The \code{fair_multiple_arrow_plot} provides an overview of the same relationships but for all possible permutations of the sensitive attributes, enabling a prioritization of different approaches. The \code{fair_density_plot} method presents the probability distributions of predictions relative to the values of the sensitive attributes. Lastly, \code{fair_waterfall_plot} demonstrates the sequential gain in fairness using waterfall plots. All of the \code{graphs} utilities also permit plots using approximate fairness, as explained in Section~\ref{subsec:approxfairness}, if $\boldsymbol{\varepsilon}$ is provided by the user. This is done by setting \code{epsilon}
to specific vector values in all of the aforementioned \code{graphs} functions. The next Section \ref{sec:illustrations} demonstrates all of these methods in a case study to illustrate their specific use. 

All described methods are provided with doc-string documentation and example usage within the source code. For additional and in-depth details of their use we refer to the next section or the dedicated package documentation \url{https://equilibration.github.io/equipy/}. 





\section{Illustrations: Case Study on Census Data}\label{sec:illustrations}



\subsection{Data and Basemodel}

We demonstrate the functionalities of \pkg{EquiPy} on data derived from the US Census, made available within the \pkg{Folktables} package \citep{ding2021retiring}, which enables a straightforward replication of all results. Here, we specifically consider a regression task, where we predict the log of an individuals'total income (based on the \emph{ACSIncome} task of \pkg{Folktables}). Note that the same logic also applies to binary classification tasks, with the condition that the model predictions for the \emph{scores} are obtained and not the predicted class label.

We opt for a simple basemodel to construct our predictor, based on a gradient boosted machine of the \pkg{LightGBM} implementation \citep{NIPS2017_6449f44a}, due to its proven track record for fast and accurate predictions and widespread use among the data science community. The data is split into three sets (\code{train}, \code{calib}, \code{test}), where for the \code{calib} set, no labels are needed. In practice, if there is no possibility to obtain a calibration set, the \code{train} data set may be used, although it is not recommended as described in~\citep{denis2021fairness}. The tuning of the model itself is considered out of the scope of the package, and we largely omit a discussion thereof, but the interested reader can find all information in the \textit{demo} section of the github repository. We suppose that all relevant hyperparameters have been gathered in a dictionary called \code{optimized_parameters} and the base model is then simply fitted and predictions are obtained as:
\begin{verbatim}   
    train_data = lightgbm.Dataset(data = X_train, label = y_train)
    model = lightgbm.train(train_set = train_data,
        params = optimized_parameters)

    predictions_calib = model.predict(X_calib)
    predictions_test = model.predict(X_test)           
\end{verbatim}

\subsection{Unfairness Mitigation}

\begin{figure}[h!]
    \centering
    \includegraphics[width=0.75\linewidth]{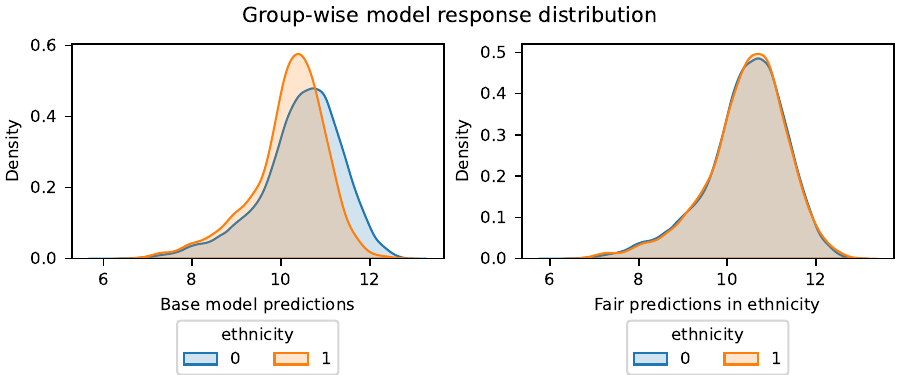}
    \caption{Group-wise model response distribution. The fair model predictions exhibit no variations across different groups.}
    \label{fig:real_density_plot}
\end{figure}

We initially investigate the prediction for a sensitive ethnicity, indexed as 1 if a given observation is a member of this group and 0 otherwise. The left pane of Figure \ref{fig:real_density_plot} depicts the densities of the initial predictions, as obtained directly from the model. Visually, there is a discrepancy between the predictions, and members of the sensitive group (indexed as 1) are predicted to have significantly lower incomes. We then proceed to measure the unfairness to quantify the bias, by using the sensitive variable \code{sens_ethn_test}:
\begin{verbatim}
    from equipy.metrics import unfairness

    unfairness(predictions_test, sens_ethn_test)
    >>> 0.437
\end{verbatim}
There seems to be a significant bias also according to the quantitative measure. The next step is to DP-calibrate the scores and post-process the test predictions. This is simply an application of the \code{FairWasserstein} function, because here we consider unfairness regarding a single sensitive attribute:
\begin{verbatim}
    from equipy.fairness import FairWasserstein

    calibrator = FairWasserstein()
    calibrator.fit(predictions_calib, sens_ethn_calib)

    predictions_test_fair = calibrator.transform(predictions_test,
        sens_ethn_test)

    unfairness(predictions_test_fair, sens_ethn_test)
    >>> 0.067
\end{verbatim} 
The simple approach reduced the unfairness significantly. To evaluate the impact on the predictive performance, we can then use the provided \code{performance} utility function:
\begin{verbatim}
    from equipy.metrics import performance

    performance(y_test, predictions_test)
    >>> 0.544

    performance(y_test, predictions_test_fair)
    >>> 0.552
\end{verbatim}
where we can detect a marginal increase in the mean squared error of the predictions, which is the default performance function. The changes can then easily be visualized using the plotting function \code{fair_density_plot} as depicted in Figure \ref{fig:real_density_plot}.
\begin{verbatim}
    from equipy.graphs import fair_density_plot

    fair_density_plot(sens_ethn_calib, sens_ethn_test, predictions_calib,
        predictions_test);
\end{verbatim}

As anticipated, based on predictions from the test set, the fair middle-ground distribution, derived from the Wasserstein barycenter, results in a perfect alignment of the conditional distributions. Consequently, the unfairness measure $\mathcal{U}_1$ is significantly reduced. 

\subsection{Unfairness Mitigation with MSA}

\begin{figure}[h!]
    \centering
    \includegraphics[width=\linewidth]{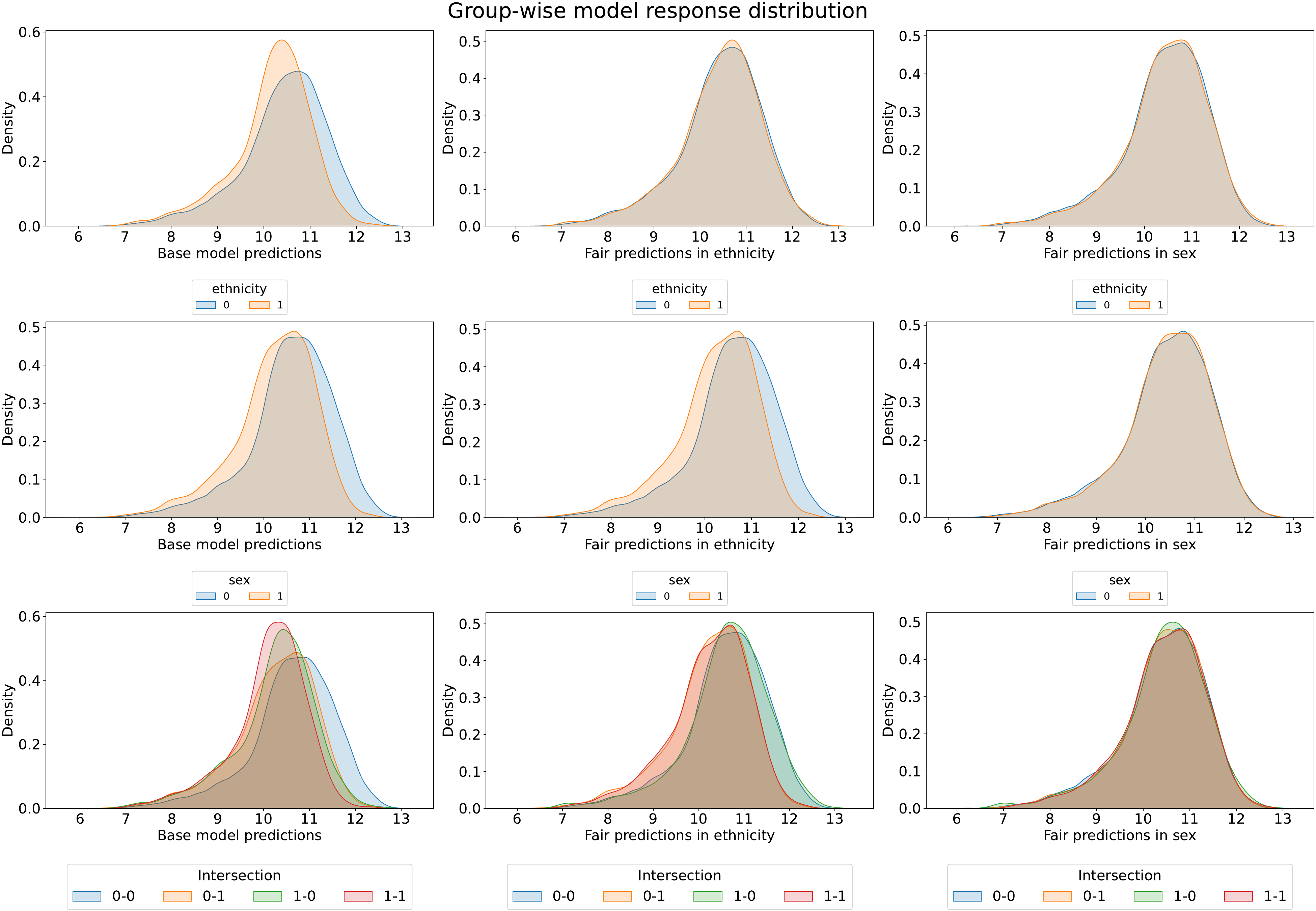}
    \caption{Group-wise model response distribution for multiple sensitive attributes: ethnicity and sex. The fair model predictions exhibit no variations across different groups.}
    \label{fig:multi_real_density_plot}
\end{figure}

The extension to the MSA case is also straightforward. As is often the case, predictions can be significantly biased with respect to more than one attribute. For example, the income distribution is often also skewed with respect to sex. Here, the predictions where fairness is imposed with respect to ethnicity can still exhibit a bias with respect to sex, or in some cases even exacerbate certain discrepancies, see \citep{hu2023sequentially} for an example thereof. \pkg{EquiPy} can handle such situations through the usage of the \code{MultiWasserstein} calibrator. Here, we specify the correction order, first addressing ethnicity and then sex. The application is similar to the SSA case:
\begin{verbatim}
    from equipy.fairness import MultiWasserstein

    sens_twovar_calib = pd.DataFrame({'ethnicity': sens_ethn_calib, 
        'sex': sens_sex_calib})
    sens_twovar_test = pd.DataFrame({'ethnicity': sens_ethn_test, 
        'sex': sens_sex_test})
                                       
    unfairness(predictions_test, sens_twovar_test)
    >>> 0.783

    calibrator_twovar = MultiWasserstein()
    calibrator_twovar.fit(predictions_calib, sens_twovar_calib)

    predictions_twovar_test_fair = calibrator_twovar.transform(
        predictions_test, sens_twovar_test)

    unfairness(predictions_twovar_test_fair, sens_twovar_test)
    >>> 0.106
\end{verbatim}
Again, the unfairness is significantly reduced. For multiple sensitive features the plotting utilities become paramount to compare the impact that each of the variables had on the final outcome. Further, the plotting utilities can help a decision maker to prioritize certain constraints over others, by providing a full impact analysis. 

\subsection{Visualization Using the Graphs Module}

\begin{figure}[h!]
    \centering
    \includegraphics[width=0.65\linewidth]{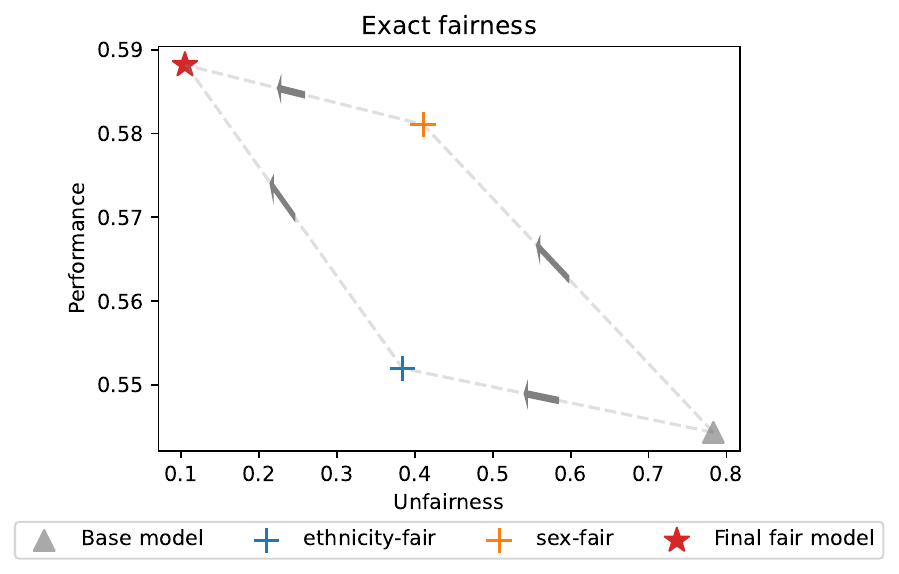}
    \caption{Arrow graphic depicting the performance-unfairness trade-off given the two sensitive features.}
    \label{fig:arrowplot}
\end{figure}

The simplest way to visualize the unfairness-performance trade-off is to use\\ \code{fair_multiple_arrow_plot}, which plots the trade-off in terms of performance (that is, mean squared error for this application) and unfairness for all possible order of correction for the sensitive attributes. This enables verification of the algorithm’s results, as the predictions should align after the final correction, regardless of the correction sequence.
To be able to compare performance on test set before and after applying the mitigation technique, \code{y_test} must be specified, which contained the true labels. Figure \ref{fig:arrowplot} can be produced by calling the graphs module as:
\begin{verbatim}
    from equipy.graphs import fair_multiple_arrow_plot
    fair_multiple_arrow_plot(sens_twovar_calib, sens_twovar_test,
        predictions_calib, predictions_test, y_test)                   
\end{verbatim}
The results show that correcting the scores for sex has a larger impact on the predictive performance as compared to correcting the scores with ethnicity first, while having a similar impact on the reduced unfairness, depicted with the higher slope for the former. This illustration enables to bring to light differences in correction strategies, in turn mitigating concerns about \emph{fairwashing} (refer to the article \cite{aivodji2019fairwashing} for a more in-depth discussion). 

\begin{figure}[h!]
    \centering
    \includegraphics[width=0.5\linewidth]{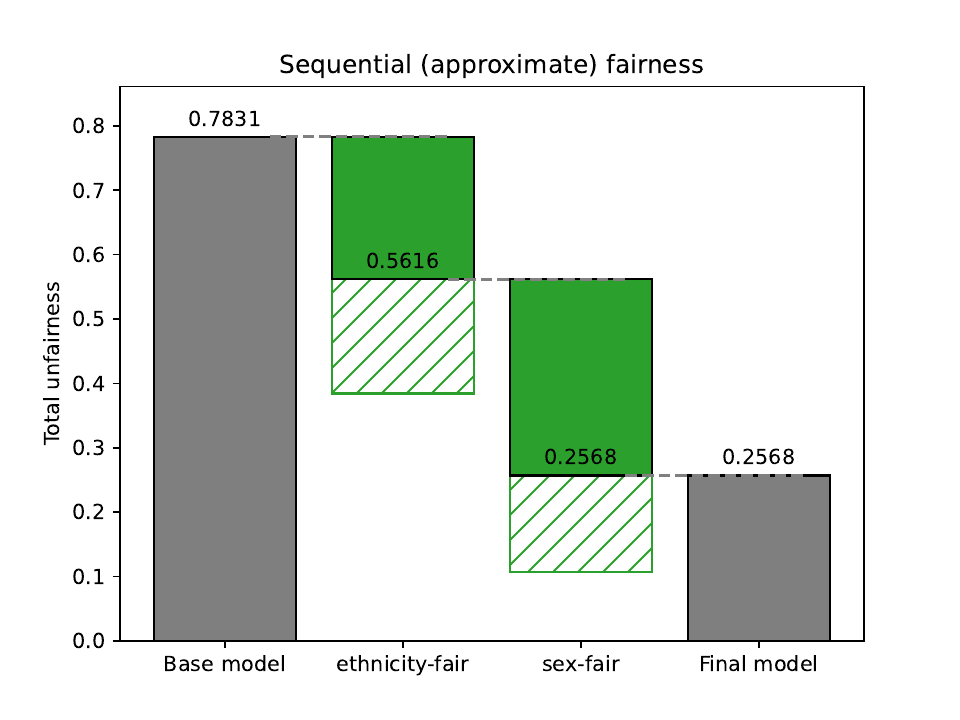}~\includegraphics[width=0.5\linewidth]{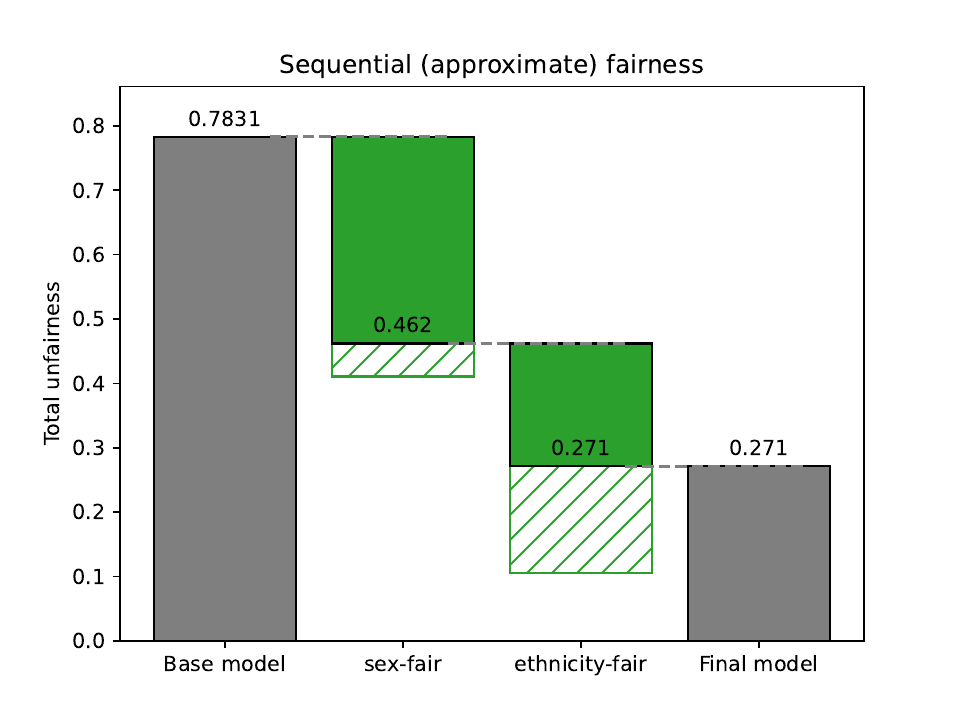}
    \caption{Waterfall plots for the two correction approaches. Left pane, correcting first for ethnicity with a $\varepsilon$ factor of 0.5 and then for sex with a $\varepsilon$ factor of 0.25. Right pane, correcting first for sex with a $\varepsilon$ factor of 0.25 and then for ethnicity with a $\varepsilon$ factor of 0.5.}
    \label{fig:waterfall}
\end{figure}

As discussed in Section \ref{subsec:approxfairness}, splitting the MSA problem into sub-problems has the additional advantage that each sensitive feature can be weighted to achieve approximate fairness. This can be visualized using the \code{fair_waterfall_plot} using a command such as:
\begin{verbatim}
    from equipy.graphs import fair_waterfall_plot
    
    fair_waterfall_plot(sens_twovar_calib, sens_twovar_test, 
        predictions_calib, predictions_test, epsilon=[0.5, 0.25])
\end{verbatim}
The goal of the waterfall plot is to visualize how the unfairness score is reduced at each step. Figure \ref{fig:waterfall} depicts the gain in fairness while applying different sequential orders for ethnicity and sex with similar values for $\boldsymbol{\varepsilon}$ in the \code{fair_waterfall_plot}. Here, correcting first for ethnicity with an epsilon factor of 50\% enables the correction of 28\% of the total unfairness (while reducing unfairness in ethnicity by 50\%) whereas correcting secondly for ethnicity with the same epsilon factor enables the correction of 24\% of the total unfairness. Given that sensitive variables can also correlate, an approximate correction might have influences in the mitigation step of the second variable. Nevertheless, the final unfairness value remains the same for both orders.


\subsection{Contribution of Each Sensitive Variable to Unfairness}

When applying the mitigation approach to MSA, it is also possible to decompose the unfairness attributed to each sensitive attribute along the sequential correction path for a given order (here, ethnicity then sex). This can be done by retrieving the dictionary \code{y_fair} from the \code{MultiWasserstein} module, which contains the initial predictions from the base model, the predictions after correcting for ethnicity in the first step, and the predictions after correcting for sex in the second and final step. The complete replication code for calculating unfairness in both ethnicity and sex at each correction step is provided in Appendix~\ref{appendix:decompo}. Table~\ref{tab:decompo} presents the unfairness values for this illustrative example.

\begin{verbatim}
    calibrator_twovar = MultiWasserstein()
    calibrator_twovar.fit(predictions_calib, sens_twovar_calib)
    
    predictions_twovar_test_fair = calibrator_twovar.transform(
        predictions_test, sens_twovar_test)
    y_seq_fair = calibrator_msa.y_fair
    
    # Unfairness in ethnicity before mitigation
     unfairness(y_seq_fair['Base model'],
        sens_twovar_test[['ethnicity']])
    >>> 0.437

    # Unfairness in sex before mitigation
     unfairness(y_seq_fair['Base model'], sens_twovar_test[['sex']])
    >>> 0.347 
\end{verbatim}

\begin{table}[t]
    \centering
    \begin{tabular}{l c c c}
        \toprule
         & $\mathcal{U}_{\text{ethnicity}}$ & $\mathcal{U}_{\text{sex}}$ & $\mathcal{U}_{\text{sex,ethnicity}}$\\
        \midrule
        Base model $f$ & 0.4366 & 0.3465 & 0.7831 \\
        1. $f_{B_{\text{ethnicity}}}$  & 0.0466 & 0.3376 & 0.3842 \\
        2. $f_B = f_{B_{\text{ethnicity,sex}}}$ & 0.0726 & 0.0338 & 0.1064 \\
        \bottomrule
    \end{tabular}
    \caption{Unfairness decomposition for sex and ethnicity calculated on Census Data. $f_{B_{\text{ethnicity}}}$ represents the model fair with respect to ethnicity, obtained after the first correction step in \pkg{EquiPy}, while $f_B$ corresponds to the model fair with respect to both ethnicity and sex, obtained after the second correction step in \pkg{EquiPy}.}
    \label{tab:decompo}
\end{table}

Unfairness decomposition in Table~\ref{tab:decompo} helps assess and interpret the contribution of each sensitive attribute to the total unfairness metric. Notably, as can be seen in the right-middle graph of Figure~\ref{fig:multi_real_density_plot}, correcting for sex in the second step \emph{almost} does not increase unfairness in ethnicity (compared to the unfairness initial value), which is desirable, as the goal is to maintain fairness across all sensitive attributes throughout the sequential correction process. One might also want to evaluate the unfairness in ethnicity after mitigating only for sex, as these two variables can be correlated. Indeed, in practice, ethnicity is often unavailable to practitioners, so it is important to ensure that ensuring fairness for one observed sensitive attribute (for example, sex) does not inadvertently compromise fairness for another unobserved attribute (for example, ethnicity).

\section{Summary and Discussion} \label{sec:summary}

The \pkg{EquiPy} package provides simple and high level functionalities to achieve fairness as defined by the Demographic Parity condition. By relying on a post-processing approach, it remains truly agnostic to the underlying models predictions that should be corrected. Instead of addressing the issue arising from multiple sensitive attributes by regrouping features, \pkg{EquiPy} relies on the associativity of Wasserstein Barycenters to achieve a sequential fairness calibration. This in turn allows a more robust estimation and also an easier interpretation of the results. Next to the mitigation techniques, the package also provides extensive visualization utilities that enable non-technical users a broader understanding of the trade-off involved.

\appendix





\clearpage

\section{More technical details} \label{app:technical}

Consider two ($k=2$) distributions, conditionnaly Gaussian, $\mathcal{N}(\mu_k,\sigma_k^2)$, Wasserstein barycenter is a Gaussian distribution $\mathcal{N}(\mu,\sigma^2)$ where
$$
\begin{cases}
\mu = w_1 \mu_1+w_2 \mu_2\\
\sigma^2 = (w_1 \sigma_1+w_2 \sigma_2)^2
\end{cases}
$$
Recall that the mixture has also mean $\mu = w_1 \mu_1+w_2 \mu_2$, but has variance
$$
v=w_1(\mu_1^2+\sigma_1^2)+
w_2(\mu_2^2+\sigma_2^2) - 
\big(w_1 \mu_1+w_2 \mu_2\big)^2 = w_1\sigma_1^2 + w_2\sigma_2^2 + w_1w_2(\mu_1 - \mu_2)^2. 
$$
Since the function $x\to x^2$ is convex, we can apply Jensen’s inequality to $\sigma^2$,
$$
(w_1 \sigma_1+w_2 \sigma_2)^2 \leq w_1 \sigma_1^2+w_2 \sigma_2^2
$$
We recognize on the right the first part of the variance of the mixture, and the second part, $w_1w_2(\mu_1 - \mu_2)^2$ is positive, therefore 
$$
\sigma^2 = 
(w_1 \sigma_1+w_2 \sigma_2)^2 \leq w_1 \sigma_1^2+w_2 \sigma_2^2+w_1w_2(\mu_1 - \mu_2)^2 =v
$$

\section{Illustrations: Unfairness decomposition} \label{appendix:decompo}

Below is the replication code from the \textit{demo} section of the github repository, which calculates unfairness for each sensitive attribute considered in the mitigation approach for multiple sensitive attributes using the \proglang{Python} package \pkg{EquiPy}, applied to the Census Data case study in Section~\ref{sec:illustrations}.

\begin{verbatim}
    calibrator_twovar = MultiWasserstein()
    calibrator_twovar.fit(predictions_calib, sens_twovar_calib)
    
    predictions_twovar_test_fair = calibrator_twovar.transform(
        predictions_test, sens_twovar_test)
    y_seq_fair = calibrator_msa.y_fair
    
    # Unfairness in ethnicity before mitigation
     unfairness(y_seq_fair['Base model'], 
        sens_twovar_test[['ethnicity']])
    >>> 0.437

    # Unfairness in sex before mitigation
     unfairness(y_seq_fair['Base model'], sens_twovar_test[['sex']])
    >>> 0.347

    # Unfairness in ethnicity after mitigation w.r.t. ethnicity
    unfairness(y_seq_fair['ethnicity'], sens_twovar_test[['ethnicity']])
    >>> 0.047

    # Unfairness in sex after mitigation w.r.t. ethnicity
    unfairness(y_seq_fair['ethnicity'], sens_twovar_test[['sex']])
    >>> 0.338

    # Unfairness in ethnicity after mitigation w.r.t. ethnicity and sex
    unfairness(y_seq_fair['ethnicity'], sens_twovar_test[['sex']])
    >>> 0.073

    # Unfairness in sex after mitigation w.r.t. ethnicity and sex
    unfairness(y_seq_fair['sex'], sens_twovar_test[['sex']])
    >>> 0.034
\end{verbatim}


    


\bibliography{refs}

\end{document}